\documentclass{article}

\usepackage{arxiv}
\usepackage[utf8]{inputenc} % allow utf-8 input
\usepackage[T1]{fontenc}    % use 8-bit T1 fonts
\usepackage{hyperref}       % hyperlinks
\usepackage{url}            % simple URL typesetting
\usepackage{booktabs}       % professional-quality tables
\usepackage{amsfonts}       % blackboard math symbols
\usepackage{nicefrac}       % compact symbols for 1/2, etc.
\usepackage{amsmath}
\usepackage[ruled,linesnumbered]{algorithm2e}
\usepackage{graphicx}
\usepackage{subfig}
\usepackage{epstopdf} 
\usepackage{tikz}
\usetikzlibrary{shapes,arrows}

\DeclareMathOperator*{\argmax}{arg\,max}
\DeclareMathOperator*{\argmin}{arg\,min}

\graphicspath{{Figs/}}

\title{Unsupervised Deep-Learning Based Deformable Image Registration: A Bayesian Framework}
\author{%
Samah~Khawaled\\
Department of Applied Mathematics\\
Technion – Israel Institute of Technology\\
Haifa, Isreal \\
\texttt{ssamahkh@campus.technion.ac.il} 
\And
Moti~Freiman\thanks{Taub fellow (supported by the Taub Family Foundation)}\\
Department of Biomedical Engineering \\
Technion – Israel Institute of Technology \\
Haifa, Isreal \\
\texttt{moti.freiman@technion.ac.il} 
}

\begin{document}

\maketitle

\begin{abstract}
	Unsupervised deep-learning (DL) models were recently proposed for deformable image registration tasks.  In such models, a neural-network is trained to predict the best deformation field by minimizing some dissimilarity function between the moving and the target images. After training on a dataset without reference deformation fields available, such a model can be used to rapidly predict the deformation field between newly seen moving and target images. Currently, the training process effectively provides a point-estimate of the network weights rather than characterizing their entire posterior distribution. This may result in a potential over-fitting which may yield sub-optimal results at inference phase, especially for small-size datasets, frequently present in the medical imaging domain. We introduce a fully Bayesian framework for unsupervised DL-based deformable image registration. Our method provides a principled way to characterize the true posterior distribution, thus, avoiding potential over-fitting. We used stochastic gradient Langevin dynamics (SGLD) to conduct the posterior sampling, which is both theoretically well-founded and computationally efficient. We demonstrated the added-value of our Basyesian unsupervised DL-based registration framework on the MNIST and brain MRI (MGH10) datasets in comparison to the \texttt{VoxelMorph} unsupervised DL-based image registration framework. Our experiments show that our approach provided better estimates of the deformation field by means of improved mean-squared-error ($0.0063$ vs. $0.0065$) and Dice coefficient ($0.73$ vs. $0.71$) for the MNIST and the MGH10 datasets respectively. Further, our approach provides an estimate of the uncertainty in the deformation-field by characterizing the true posterior distribution.
\end{abstract}

\section{Introduction}
Deformable image registration is a fundamental task needed in a wide-range of computer vision and image analysis applications. In particular, it plays a key role in the medical imagining domain, where the execution of pre-alignment of 2D/3D voxels is crucial for many medical applications: motion compensation, multi-modal analysis, \textit{inter} or \textit{intra} subject alignment for change detection and longitudinal analysis, to name a few \cite{hill2001medical,zitova2003image}. For a comprehensive overview of both classical and deep-learning (DL) based registration methods and their different categories we refer the reader to \cite{sotiras2013deformable} and \cite{haskins2020deep}. 

Classical methods formulate the ill-posed deformable registration task as a regularized optimization problem over the deformation fields space and solve it using iterative solvers \cite{ashburner2007fast,bajcsy1989multiresolution,avants2008symmetric}. Commonly used regularization terms encourage smoothness of the local deformations and ensures desirable properties such as topology-preserving fields \cite{beg2005computing, ashburner2007fast, avants2008symmetric}. However, these conventional algorithms are computationally demanding, which, in turn, makes the registration of a new pair of images a computationally expensive process. 

In light of the success of DL-based methods in numerous computer vision tasks, several studies aimed to propose more efficient and less time-consuming deformable registration approaches based on DL models \cite{sandkuhler2019recurrent,dalca2019learning,de2019deep,balakrishnan2018unsupervised,dalca2019varreg,dalca2018varreg,yang2017quicksilver,krebs2019learning}. These techniques learn a deformation field prediction model either through a supervised learning framework (i.e with the help of provided reference data) \cite{yang2017quicksilver} or in unsupervised manner \cite{de2019deep,balakrishnan2018unsupervised,dalca2018varreg,krebs2019learning}. The latter is advantageous as it doesn't require available reference deformation fields. 

However, the current training process effectively provides point-estimates of the neural-network weights rather than characterizing their entire posterior distribution. This may result in a potential over-fitting that yields sub-optimal results at inference phase, especially for small-size datasets commonly present in the medical imaging domain. Further, it lacks the ability to quantify the uncertainty of the deformation field parameter estimates. 

Bayesian learning in DL models plays a critical role in improving generalization and assessing the uncertainty of the predictions  \cite{cheng2019bayesian,mackay1992practical,neal2012bayesian}. 
In the context of image registration, Yang et al. \cite{yang2017quicksilver} used an inference-time dropout layer to extend deformation field point-estimates obtained with their \texttt{QuickSilver} supervised registration predictive model to a probabilistic model, enabling the assessment of the uncertainty of the predicted deformation-field. Dalca et al. \cite{dalca2018varreg, dalca2019varreg} aimed to characterize the posterior distribution by modeling the latent variable that parameterizes the deformation field with mean and variance parameters in their unsupervised \texttt{VoxelMorph} framework \cite{balakrishnan2018unsupervised}. They introduced a Kullback-Leibler (KL) Divergence term to their loss function to encourage the network prediction toward a normal distribution. A similar method was used by Krebs et al. \cite{krebs2019learning} for cardiac MRI registration. This approach assumes a normally distributed posterior which may result in sub-optimal results in cases of other posterior distributions. While these approaches do provide some mechanism to quantify uncertainty in the deformation field prediction, they do not provide a fully Bayesian characterization of the deformation field posterior distribution. 

In a shift from previous approaches, we propose a new, non-parametric, Bayesian unsupervised image registration framework. Our approach does not make any assumptions on the actual posterior distribution. Therefore has the potential to improve upon parametric methods in case of non-normal posterior. We adopt the strategy of stochastic gradient Langevin dynamics (SGLD) \cite{welling2011bayesian} to perform the sampling from the posterior distribution of the network weights \cite{cheng2019bayesian}. 

SGLD-based training is computationally efficient and suits a wide-range of Bayesian estimation themes encountered in image-processing tasks. For example, Cheng et al. \cite{cheng2019bayesian}, show considerable improvements in unsupervised image denoising by using SGLD to perform posterior sampling in a deep-image-prior denoising model \cite{ulyanov2018deep}. Further, noise injection to the weights' gradients has shown additional benefits in training deep and complex models beyond enabling efficient posterior sampling \cite{neelakantan2015adding}. Some of these are better solutions for non-convex problems, improved generalization ability, and reduced over-fitting, among others. 
%Noise gradient scheduling can be done by means of the standard SGLD-based strategies or by incorporating noise to other alternative stochastic optimization techniques \cite{welling2011bayesian}. 
Recently, various studies proposed a version of SGLD that combines both Gaussian noise and \textit{adaptive} optimization algorithms \cite{chen2016bridging,neelakantan2015adding}. 
%The Santa algorithm, for example, bridges the gap between these two landscapes and extends the SGLD-MCMC methods to an adaptive version. 

Specifically, during the training phase of our framework, we inject Gaussian noise with \textit{adaptive} variance to the loss gradients and keep all weights obtained after the \textit{"burn-in"} iteration.\footnote[1]{Following the notation of SGLD \cite{welling2011bayesian} and that used in \cite{cheng2019bayesian}, we use the term \textit{"burn-in"} to denote the iteration (epoch) where the loss is stable and converges to its minima. The behavior of the loss at this region is dominated by small variations caused by the injected noise.}
We estimate the posterior distribution of the deformation field during inference by averaging deformation field predictions obtained by the model with the saved weights which, in turn, provide robustness to noisy images. In addition, we quantify the uncertainty in the predicted deformation field by calculating the empirical variance of the samples. Our predictive regression model is built upon the \texttt{VoxelMorph} registration system \cite{balakrishnan2018unsupervised}. We use the same network architecture, while incorporating our Bayesian strategy, in both the training and the inference phases. We demonstrated the added-value of our approach on 2D unsupervised registration of both the MNIST \cite{lecun1998mnist} and a brain MRI (MGH10) \cite{klein2009evaluation} databases.             

\section{Background: Image Registration}\label{sec:back}
The deformable registration task can be formulated as an optimization problem. Let us denote the pair of fixed and moving images by $I_{f}$ and $I_M$, respectively. $\Phi$ is the deformation field, which accounts for mapping the grid of $I_M$ to the grid of $I_f$. Then, the energy functional that we aim to optimize is:
\begin{equation}
\argmin_\Phi {S(I_f,I_M\circ\Phi)+\lambda R(\Phi)} \label{eq:1}
\end{equation}
where $I_M\circ\Phi$ denotes the result of warping the moving image with $\Phi$. $S$ is a dissimilarity term, which quantifies the resemblance between the resulted image and the fixed input, and $R$ is a regularization term that encourages the deformation smoothness. The scalar $\lambda$ is a tuning-parameter that accounts for balancing between the two terms, and it controls the smoothness of the resulted deformation. Popular dissimilarity metrics are the mean squared difference/error (MSE), the cross-correlation and the mutual information. The former is used when images have the same dynamic range of intensities while the latter are essential when images exhibit varying gray-levels and different contrasts.\\ 
\indent In DL-based registration, the task of the deep model is to predict the deformation:
\begin{equation}
\hat{\Phi} = f_\theta(I_{M},I_{F}) \label{eq:2}
\end{equation} 
where $\theta$ are the parameters of the network and $I_{M},I_{F}$ are the input pairs. It should be emphasized that, in contrast to supervised schemes, unsupervised registration models "learns" the deformation field that maps from $I_M$ to $I_F$ without providing reference transformation. In the case of non-parametric (free-form) deformation,\footnote[2]{Here, we don't consider the parametric or model-based deformation (such as an affine-based or a spline-based) and we limit the discussion to the free-form only. In the parametric case, the network predicts the parameters of the deformation.} it predicts $\hat{\Phi}$ by optimizing the following: 
\begin{equation}
\hat{\theta}= \argmin_\theta {S(I_F,I_M\circ f_\theta(I_{M},I_{F}))+\lambda R(f_\theta(I_{M},I_{F}))} \label{eq:3}
\end{equation}   

\section{The Bayesian Approach}\label{sec:approach}
From a probabilistic point of view, the network training aims to maximize the posterior estimation of the deformation field parameters:
\begin{equation}
\hat{\theta}= \argmax_\theta {P\left(\theta|I_F,I_M\right) \propto P(I_F,I_M|\theta)P(\theta)} \label{eq:4}
\end{equation} 
Since direct integration of the posterior distribution is intractable, several approaches proposed to achieve a maximum posterior estimation in a computationally feasible time. The \texttt{VoxelMorph} approach \cite{balakrishnan2018unsupervised} introduced a smoothness regularization term to constrain the deformation field parameters rather than explicitly define the distribution of the deformation field parameters. While this can provide a maximum-a-posterior (MAP) point estimation of the deformation field parameters, it cannot characterize the entire distribution of the deformation field parameters, which may lead to sub-optimal results. In a more recent work, the \texttt{VoxelMorph} approach was extended to provide a characterization of the uncertainty in the deformation field parameters by assuming that the latent variable that parameterizes the deformation field is distributed normally with zero mean and some variance \cite{dalca2018varreg,dalca2019varreg}. During the learning process a KL divergence term is introduced to the loss function to encourage the network prediction towards a normal distribution. However, this approach assumes a normally distributed posterior which may result in sub-optimal results in case of other posterior distributions.

Instead, we efficiently sample the actual posterior distribution of the model weights using an {\it adaptive} SGLD mechanism. We treat the network weights as random variables and aim to sample the posterior distribution of the model prediction. To this end, we incorporate a noise scheduler that injects a time-dependent Gaussian noise to the gradients of the loss during the optimization process. At every training iteration, we add Gaussian noise with \textit{adaptive} variance to the loss gradients. Then, the weights are updated in the next iteration according to the "noisy" gradients. This noise schedule can be performed with any stochastic optimization algorithm during the training procedure. In this work we focused on the formulation of the method for the \texttt{Adam} optimizer. We use Gaussian noise with a variance proportional to the learning rate of the \texttt{Adam} optimization algorithm, as it allows the adaption of the noise to the nature of loss curve. Moreover, previous research shows better performance using noise adaptive or time-dependent variance than constant variance \cite{neelakantan2015adding}. Our {\it adaptive} SGLD-based registration (ASGLDReg) strategy is outlined in algorithm \ref{alg:1}. The proposed approach can be directly extended to other stochastic optimization algorithms as well.
\begin{algorithm}[t] 
	\SetAlgoLined
	\SetKwFunction{Ftrain}{TrainNetwork}
	\SetKwFunction{Ftest}{FeedForward}
	\caption{\texttt{ASGLD}Reg Algorithm}\label{alg:1}
	\SetKwInput{KwInput}{Input}                % Set the Input
	\SetKwInput{KwOutput}{Output}
	\KwInput{Tuning parameters $\lambda$ and $\alpha$, and number of epochs $N$}
	\KwOutput{Estimated deformation $\hat{\Phi}$ and registered image $I_R$}
	\KwData{Dataset of pairs of fixed and moving images $I_M,I_F$}	
	\tcc{\underline{The \textit{offline} training}} 
	\SetKwProg{Fn}{Function}{:}{}
	\Fn{\Ftrain{${I_M,I_F}$,$\lambda$,$\alpha$,$N$}}
	{
		\For{$t<N$}
		{  
			$\tilde{g}^{t}\leftarrow g^{t}+\textbf{N}^{t}$, where $\textbf{N}^{t}\sim\mathcal{N}(0,\frac{s^{t}}{\alpha})$
			and $s^{t}$ is the \textit{adaptive} step size. \\
			$\theta^{t+1}\leftarrow \text{\textsc{Adam\_Update}}(\theta^{t},\tilde{g}^{t})$ \\

		}
		\KwRet{$\left\{\theta_{t}\right\} _{t_{b}}^{N}$} \\
	}
	\setcounter{AlgoLine}{0}
	\tcc{\underline{\textit{Online} Registration}} 
	\Fn{\Ftest{$I_M$,$I_F$}}
	{
		Compute a set of deformations $\left\{ \Phi_{t}\right\} _{t_{b}}^{N}$ by feed-forwarding $I_M$, $\Phi_{t}=f_{\theta^t}(I_M,I_F)$ for $t\in \left[t_b,N\right]$ \\
		Estimate the deformation: 
		$\hat{\Phi}=$$\frac{\sum_{t=t_b}^{N}\Phi_t}{N-t_{b}}$\\ 
		Register Image: $I_{R}\leftarrow\text{\textsc{Spatial\_Transform}}(I_{in},\hat{\Phi}).$ \\
		
		\KwRet{$I_R,\hat{\Phi}$}
	}
\end{algorithm}  
\paragraph{Offline Training}
$L(I_M,I_F,f_\theta(I_M,I_F))$ denotes the overall registration loss, described in eq.\eqref{eq:3}, which is composed of both similarity and regularization terms. We denote the loss gradients by:
\begin{equation}
g^t\overset{\triangle}{=}\nabla_{\theta}L^t(I_{M},I_{F},f_{\theta}(I_{M},I_{F}))
\end{equation}
where $t$ is the training iteration (epoch). At each training iteration, Gaussian noise with time-varying variance is added to $g$:
\begin{equation}
\tilde{g}^{t}\leftarrow g^{t}+\textbf{N}^{t} 
\end{equation}
where $\textbf{N}^{t}\sim\mathcal{N}(0,\frac{s^{t}}{\alpha})$
and $s^{t}$ is the \texttt{Adam} step size. $\alpha$ is a user-selected parameter that controls the noise variance. This is especially important in the first learning stages, which involve a large step size. 
The network parameters are then updated according to the \texttt{Adam} update rule:
\begin{equation} 
\theta^{t+1}\leftarrow \theta^{t} - s^{t}\hat{m}^{t}
\end{equation}
where $s^{t}=\frac{\eta}{\sqrt{\hat{v}^{t}+\epsilon}}$, $\hat{m}^{t}$ and $\hat{v}^{t}$ are the bias-corrected versions of the decaying averages of the past gradients and the second moment (squared gradients), respectively:
\begin{align}
\hat{m}^{t} = \nicefrac{m^{t}}{1-\beta_{1}^{t}} \nonumber \\
\hat{v}^{t} = \nicefrac{v^{t}}{1-\beta_{2}^{t}}
\end{align}
where $m^{t}=m^{t-1}+(1-\beta_{1})\tilde{g}^{t}$ and $v^{t}=v^{t-1}+(1-\beta_{2})(\tilde{g}^{t})^2$. $\beta_{1}$, $\beta_{2}$ are decay rates and $\eta$ is a fixed constant. 
%\begin{align}
%m^{t}=m^{t-1}+(1-\beta_{1})\tilde{g}^{t} \nonumber \\
%v^{t}=v^{t-1}+(1-\beta_{2})(\tilde{g}^{t})^2
%\end{align}

\paragraph{Online Registration} 
Having the network trained, and its weights during training saved, we exploit only the outputs of the network with weights obtained after the \textit{burn-in} phase. The \textit{burn-in} phase is the initial stage of SGLD learning, at which the step size is still relatively large, thus the gradients magnitude dominates the injected noise. After this phase, the step size decays, thus, the injected Gaussian noise will dominate the loss behavior as its  curve converges and exhibits only small variations around its steady-state. 
We sample a set of deformations $\left\{ \Phi_{t}\right\} _{t_{b}}^{N}$, obtained by feed-forwarding the pairs ${I_M,I_F}$ to our network after the \textit{burn-in} phase. $t_b$ denotes the cut-off point of the \textit{burn-in} phase and $N$ is the iterations number. Under some feasible constraints on the step size, the sampled weights converge to the posterior distribution \cite{welling2011bayesian}. Thus, outputs of our network after the \textit{burn-in} phase, trained with the {\it adaptive} SGLD approach can be considered as a sampling from the true posterior distribution. 
Then, when we have a new pair for alignment, we can estimate the averaged posterior deformation:
\begin{equation}
\hat{\Phi}=\frac{\sum_{t=t_b}^{N}\Phi_t}{N-t_{b}}
\end{equation} 
Lastly, we register the moving image by resampling its coordinate system with the spatial transform $\hat{\Phi}$. The function \textsc{Spatial\_Transform} performs spatial warping as described in \cite{balakrishnan2018unsupervised}. For each pixel $p$, a sub-pixel location $\hat{\Phi}(p)$ in $M$ is calculated. Then, the values are linearly interpolated to obtain an integer.  

\subsection{Output Statistics}
In \cite{cheng2019bayesian}, it is proved that a convolutional network (CNN) with random parameters, which is fed by a stationary input image (such as white noise), acts as a spatial Gaussian process with a stationary kernel. This is valid in the limit as the channels number in each layer goes to infinity. In addition, the authors analyze the network statistical behavior for models with beyond two layers or with more complex systems that incorporate down-sampling, sampling or skip connection. Similar to these conclusions, the deformation obtained by a CNN model with random parameters that operates on stationary pair of images behaves like a Gaussian field. We assume that the network parameters, $\theta$ are Gaussian. In our \textit{adaptive} learning setting, the latter is accomplished under the assumption of negligible second order moment of \texttt{Adam}, $v^t$, i.e. when $\beta_{2}\simeq1$.     

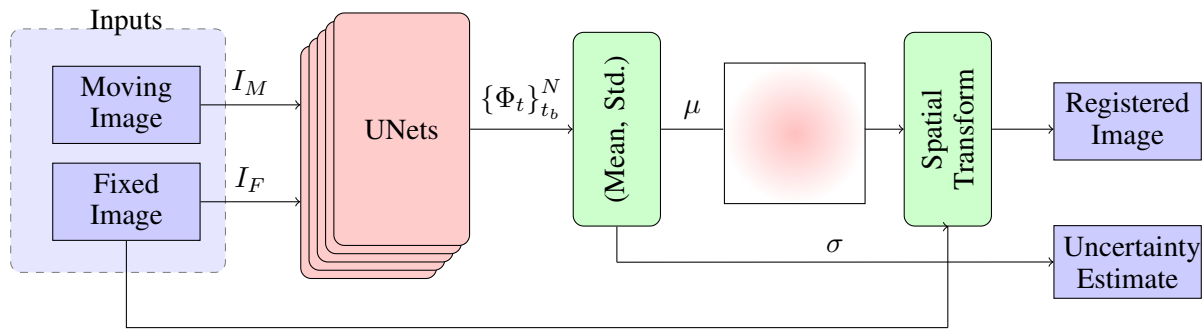
\begin{figure}[t]
	{
		\pgfdeclarelayer{background}
		\pgfdeclarelayer{foreground}
		\pgfsetlayers{background,main,foreground}
		
		% Define a few styles and constants
		\tikzstyle{sensor}=[draw, fill=blue!20, text width=4.4em, 
		text centered, minimum height=2.5em]
		\tikzstyle{ann} = [above, text width=4em]
		\tikzstyle{naveqs} = [sensor, text width=4em, fill=red!20, 
		minimum height=8em, rounded corners]
		\tikzstyle{naveqs1} = [sensor, text width=6em, fill=green!20, 
		minimum height=3em, rounded corners]
		
		\def\blockdist{2.1}
		\def\edgedist{3}
		
		\begin{tikzpicture}[scale=1.1, transform shape]
		\node (naveq) at (0,0) [naveqs] {UNet};
		\node (naveq1) at (0.1,0.1) [naveqs] {UNets};
		\node (naveq2) at (0.2,0.2) [naveqs] {UNets};
		\node (naveq3) at (0.3,0.3) [naveqs] {UNets};
		\node (naveq4) at (0.4,0.4) [naveqs] {UNets};
		\node (naveq5) at (3,0.4) [naveqs1,rotate=90] {(Mean, Std.)};
		\node (naveq6) at (7,0.4) [naveqs1,rotate=90] {Spatial Transform};
		
		% Note the use of \path instead of \node at ... below. 
		\path (naveq.140)+(-\blockdist,0) node (gyros) [sensor] {Moving Image};
		\path (naveq.-150)+(-\blockdist,0) node (accel) [sensor] {Fixed Image};
		\path (naveq.140)+(10,-0.2) node (reg) [sensor] {Registered Image};
		\path (naveq.140)+(10,-1.9) node (sigma) [sensor] {Uncertainty Estimate};
		
		\path [draw, ->] (gyros) -- node [above] {$I_M$} 
		(naveq.west |- gyros) ;
		
		\path [draw, ->] (accel) -- node [above] {$I_F$} 
		(naveq.west |- accel);
		
		\node (IMU) [above of=gyros] {Inputs};
		\path (naveq.south west)+(-0.6,-0.4) node (INS) {};

		\draw [->] (naveq4) -- node [above](\edgedist,0) {$\left\{ \Phi_{t}\right\} _{t_{b}}^{N}$} ( naveq5.north |- naveq4);

		\shadedraw[inner color=pink,outer color=white, draw=black] (4.3,-0.5) rectangle +(1.7,1.7);
		%\draw[-] (naveq5.south)++(0,0) |- (4,0.4);
		\draw (naveq5.south) -- node[above] {$\mu$} ++(0.75,0);
		\draw[->] (6,0.4) |- (naveq6.north);
        \draw[-] (naveq5.west) -- node[below] {} ++(0,-0.43);
        \draw[->] (naveq5.west)++(0,-0.43) -- node[above] {$\sigma$} (sigma.west);

		%\draw[->] (accel.south)++(0,0) |- (naveq6.west);
		\draw[-] (accel.south)++(0,0) |- (7,-2);
		\draw[->] (7,-2) |- (naveq6.west);
		\draw [->] (naveq6) -- node [above](\edgedist,0) {} (reg.west |- naveq6);
		
		\begin{pgfonlayer}{background}
		%% Compute a few helper coordinates
		\path (gyros.west |- naveq.north)+(-0.5,0.2) node (a) {};
		\path (INS.south -| naveq.east)+(+0.2,-0.2) node (b) {};
		%\path[fill=yellow!20,rounded corners, draw=black!50, dashed]
		%(a) rectangle (b);
		%\path (gyros.north west)+(-0.2,0.2) node (a) {};
		\path (gyros.north -| gyros.east)+(+0.3,-2.5) node (b) {};
		\path[fill=blue!10,rounded corners, draw=black!50, dashed]
		(a) rectangle (b);
		\end{pgfonlayer}
		
		\end{tikzpicture}
	}
	\caption{Block diagram of the proposed Bayesian registration system. After having $\left\{ \Phi_{t}\right\} _{t_{b}}^{N}$, we calculate the mean and std. of the deformation, $\mu$ and $\sigma$, respectively. Only the mean is used in the registration scheme, but, $\sigma$ provides an estimate of the result uncertainty. }
	\label{fig:bdiagram}
\end{figure}
\section{The Registration System}\label{sec:system}
Figure \ref{fig:bdiagram} illustrates the design of our Bayesian registration framework. Our main building-block is a UNet-based \cite{ronneberger2015u} CNN similar to the \texttt{VoxelMorph} model \cite{balakrishnan2018unsupervised}. Given a pair of moving ($I_M$) and target ($I_F$) images as a 2-channel input, it predicts the deformation field, $\Phi=f_\theta(I_{M},I_{F})$. Sampling from the UNet outputs after the \textit{burn-in} phase is analogous to having a set of stochastic UNets characterized by different weights, each operates on the same pair of moving and fixed images and models the corresponding deformation field. The operation of the system at the inference stage is as follows: it takes a pair of moving ($I_M$) and target ($I_F$) images as a 2-channel input and predicts the posterior deformation field $\hat{\Phi}$ by computing the average of the deformation field predictions obtained by the \textit{stochastic} UNets. Lastly, it maps each pixel, $p$, in the moving image to $\hat{\Phi}(p)$ by applying the spatial transform function. 

\subsection{Network Architecture}
Our main UNet-based building block is comprised from encoder and decoder with skip connections. Both encoder and decoder parts consists of CNN layers with kernel size $3\times3$ \footnote[3]{For 3D registration the kernel size can be modified to $3\times3\times3$.} followed by Leaky ReLU activation functions. The encoder has $4$ CNN layers, each of $32$ channels, whereas, the decoder consists of $6$ layers with the following number of channels:$\left\{32,32,32,32,32,16\right\}$. Afterwards, a 2-channel convolutional layer is applied on the UNet output to obtain the 2D deformation field. To achieve pyramidal representation of features, from the fine to coarser levels, strided Convolutions are used in the encoder. In the decoder, upsampling, convolutions, and skip connections are applied. Skip connections concatenate features learned in the first four encoder convolutional layers to the first three and fifth decoder layers. In this manner, the decoder output models a 2-channel deformation field of the same size as the fixed and moving images. A detailed description of the \texttt{VoxelMorph} model's architecture can be found in \cite{balakrishnan2018unsupervised}. 

\subsection{Loss Function}
The training process involves the optimization of the energy-functional described in eq.~\eqref{eq:3}. The UNet model is served as an energy optimization solver. We minimize the energy functional w.r.t the network parameters, $\theta$. We use the mean squared error (MSE) for our MNIST experiments and the negative local cross correlation (LCC) for our MRI images experiments to characterize the dissimilarity between the fixed image $I_{F}$ and that obtained after mapping, $I_{M}\circ\hat{\Phi}$. The LCC is commonly used for MRI registration tasks rather than the MSE due to its robustness to intensity changes. Other dissimilarity metrics such as the mutual information \cite{nan2020differentiable} can be used as well.  We use the $\text{L}_{2}$ norm over the deformation field gradients as a regularization term to encourage the deformation field smoothness. The UNet model then learns the optimal weights by minimizing the loss function composed of the two aforementioned terms.   

\begin{figure}[t]
	\begin{minipage}[b]{0.45\linewidth}
		\centering
		\subfloat[\label{fig2:a}\footnotesize ]{
			\includegraphics[width=5.3cm]{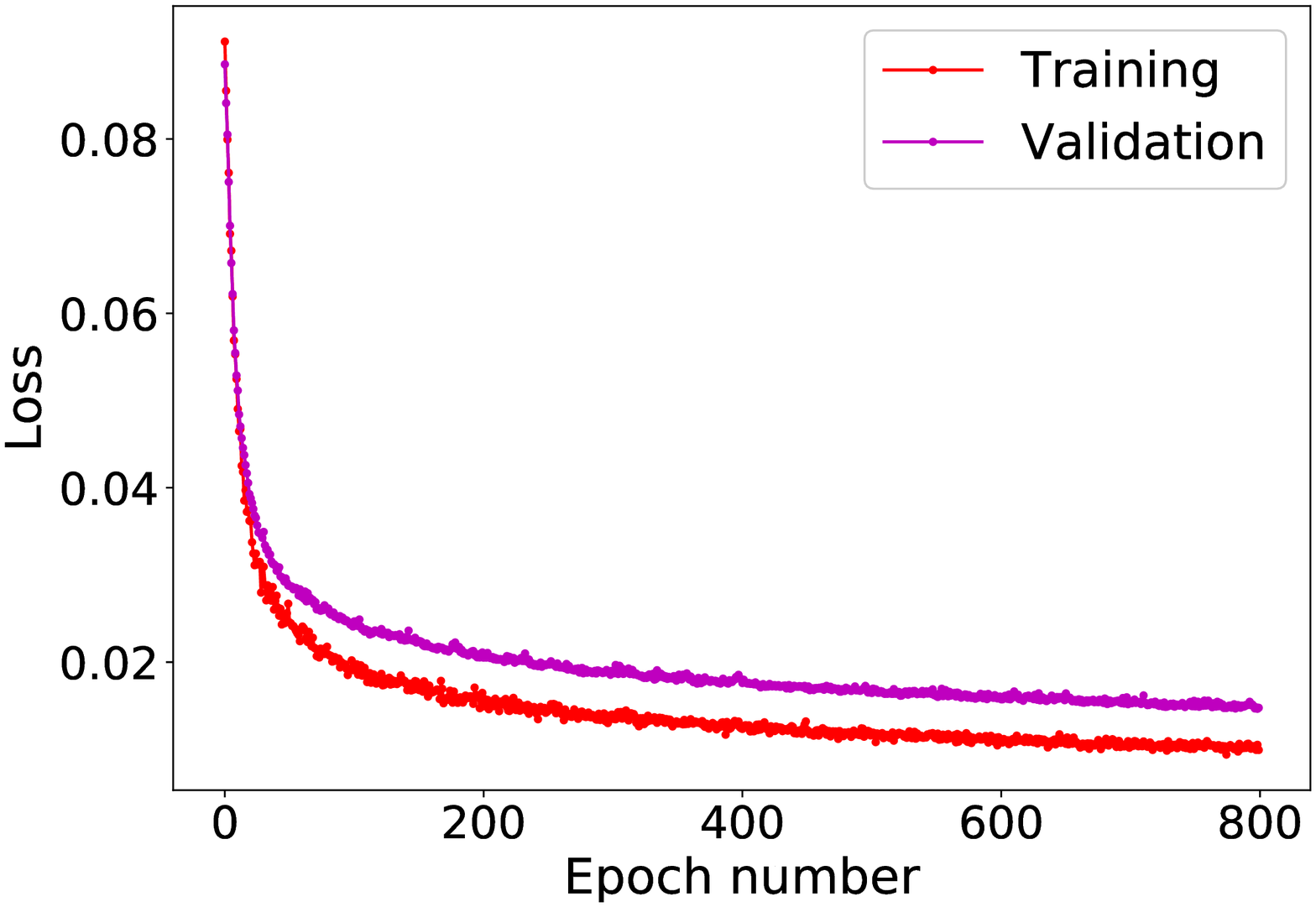}}
	\end{minipage}
	\begin{minipage}[b]{0.45\linewidth}
		\centering
		\subfloat[\label{fig2:b}\footnotesize]{
			\includegraphics[width=6.3cm]{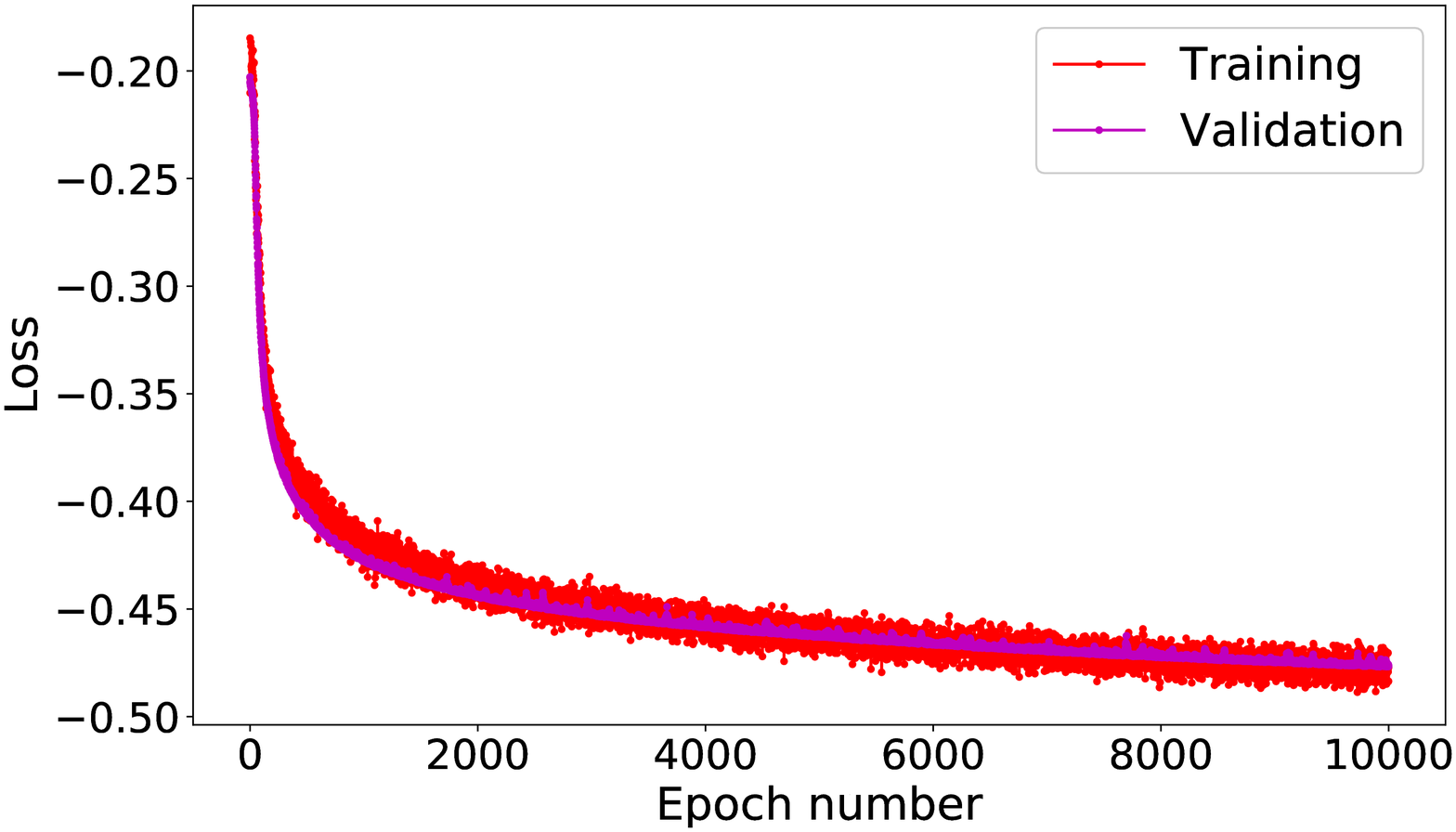}}
	\end{minipage}
	\caption{Loss curves as a function of the training iteration for both MNIST \protect\subref{fig2:a} and MGH10 \protect\subref{fig2:b} datasets. The presented loss consist of both similarity and regularization terms. In the case of MNIST, the loss shows stable behavior after $500$ iterations. However, for the MRI data it reaches a stable phase (for validation) after about $8000$ iterations.  Optimal values of the cut-off point, $t_b$, should be within these regimes. }\label{fig:loss}
\end{figure}

\section{Experiments}\label{sec:exp}
\begin{figure}[t]
    \hspace{1mm}
	\begin{minipage}[b]{0.19\linewidth}
		\centering
		\subfloat[\label{fig3:a}\footnotesize  $I_M$]{
			\includegraphics[width=2.8cm]{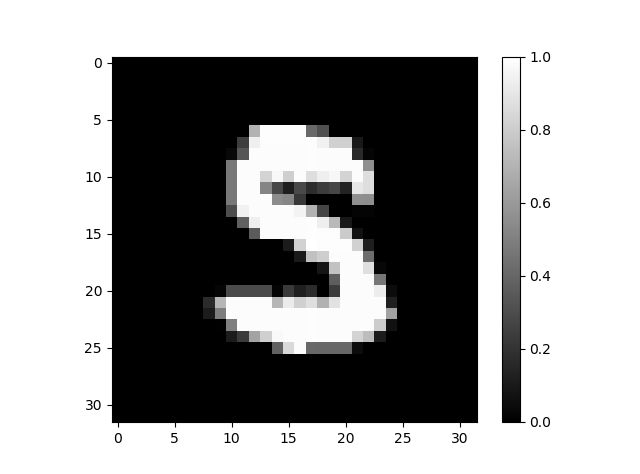}}
	\end{minipage}
	\begin{minipage}[b]{0.19\linewidth}
		\centering
		\subfloat[\label{fig3:b}\footnotesize $I_F$]{
			\includegraphics[width=2.8cm]{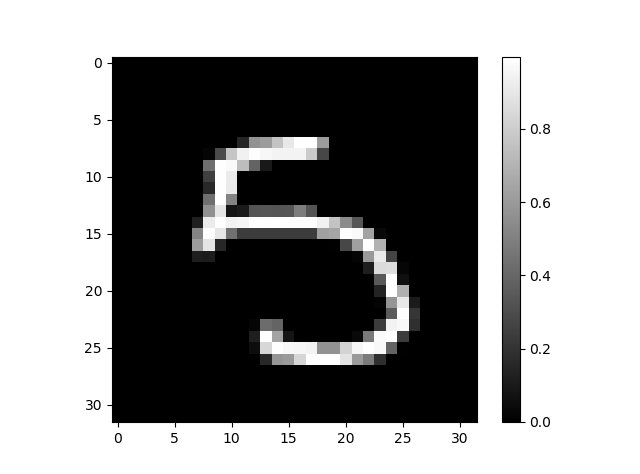}}
	\end{minipage}	
	\begin{minipage}[b]{0.19\linewidth}
		\centering
		\subfloat[\label{fig3:c}\footnotesize x-component of $\hat{\Phi}$]{
			\includegraphics[width=2.8cm]{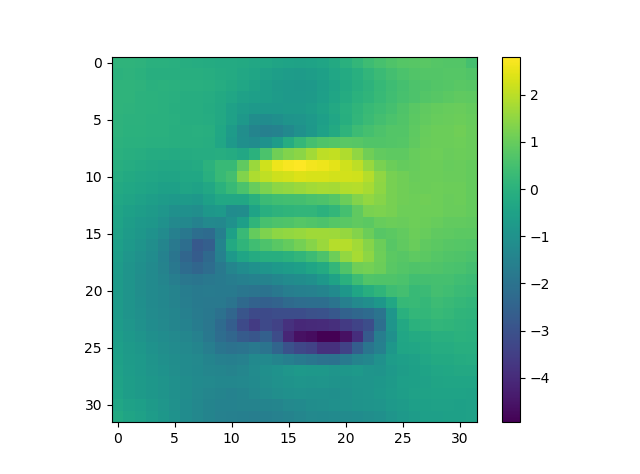}}
	\end{minipage}
	\begin{minipage}[b]{0.19\linewidth}
		\centering
		\subfloat[\label{fig3:d}\footnotesize $A_M\circ \hat{\Phi}$]{
			\includegraphics[width=2.8cm]{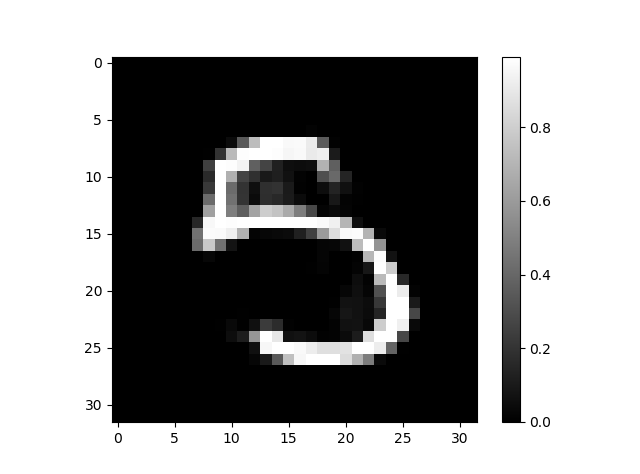}}
	\end{minipage}
	\begin{minipage}[b]{0.19\linewidth}
		\centering
		\subfloat[\label{fig3:e}\footnotesize Variance]{
			\includegraphics[width=2.8cm]{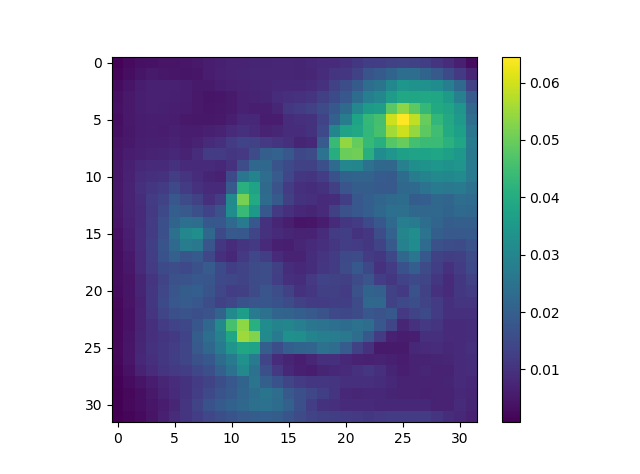}}
	\end{minipage}
	\\
	\begin{minipage}[b]{0.19\linewidth}
		\centering
		\subfloat[\label{fig3:f}\footnotesize  $I_M$]{
			\includegraphics[width=3cm]{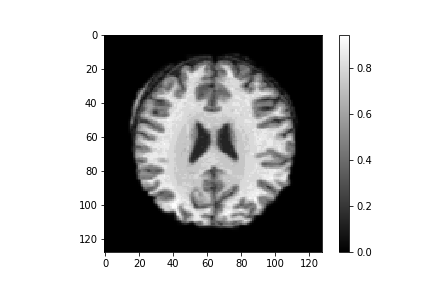}}
	\end{minipage}
	\begin{minipage}[b]{0.19\linewidth}
		\centering
		\subfloat[\label{fig3:g}\footnotesize $I_F$]{
			\includegraphics[width=3cm]{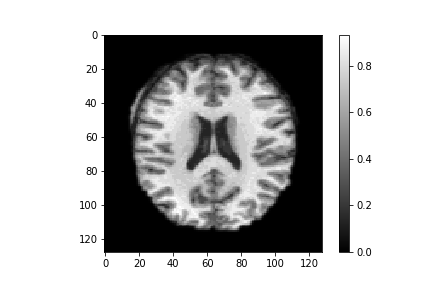}}
	\end{minipage}
	\begin{minipage}[b]{0.19\linewidth}
		\centering
		\subfloat[\label{fig3:h}\footnotesize x-component of $\hat{\Phi}$]{
			\includegraphics[width=3cm]{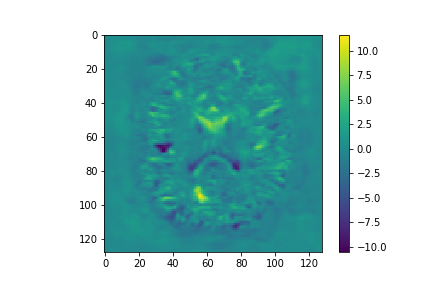}}
	\end{minipage}	
	\begin{minipage}[b]{0.19\linewidth}
		\centering
		\subfloat[\label{fig3:i}\footnotesize $A_M\circ \hat{\Phi}$]{
			\includegraphics[width=3cm]{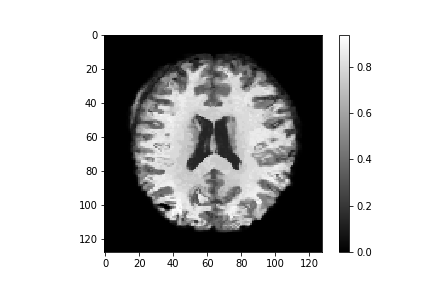}}
	\end{minipage}
	\begin{minipage}[b]{0.19\linewidth}
		\centering
		\subfloat[\label{fig3:j}\footnotesize Variance]{
			\includegraphics[width=3cm]{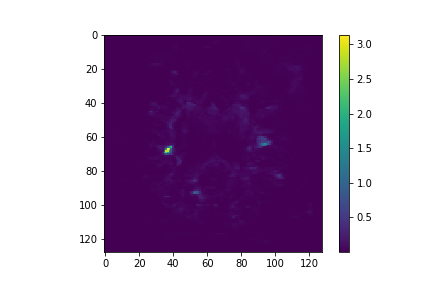}}
	\end{minipage}
	\caption{Registration Results. \protect\subref{fig3:a}, \protect\subref{fig3:f} moving and \protect\subref{fig3:b}, \protect\subref{fig3:g} fixed input images belong to MNIST and MGH10 datasets, respectively. \protect\subref{fig3:c}, \protect\subref{fig3:h} The x-component of the averaged deformation field, calculated after the \textit{burn-in}. \protect\subref{fig3:c}, \protect\subref{fig3:h} Results of registration obtained by warping the input image with the corresponding deformation field. \protect\subref{fig3:e}, \protect\subref{fig3:j} The variance of the x-component of the predicted deformation field.}\label{fig:regresults}	
\end{figure}

We implement our \texttt{ASGLD}Reg method for unsupervised 2D deformable image registration. We use the publicly available implementation of the \texttt{VoxelMorph} model as our building block.\footnote[4]{\url{https://github.com/voxelmorph/voxelmorph}} Our system was implemented using Keras with a tensorflow backend \cite{abadi2016tensorflow}. We used the MNIST \cite{lecun1998mnist} and the MGH10 brain MRI \cite{klein2009evaluation} datasetes. 
The MGH10 dataset consists of brain MRI scans of 10 subjects with provided segmentation into 74 regions. MRI images affine-registered according to the MNI152 template \cite{evans1992mri}, and preprocessed by inhomogeneity-correction. All scans are of size $182\times218\times182$ and uniform spacing grid of $1mm$ in each dimension. A detailed description of each experiment follows.
\paragraph{MNIST Registration}
We consider deformable registration of pairs of digit $5$ images, which exhibit different geometrical shapes. Data of digit $5$ was randomly split into training, validation and test sets. $199,800$ pairs were produced from $200$ examples that were selected for training, $100$ pairs for validation and $1000$ for the test set. Images were resized to $32\times 32$ and their intensities were normalized to the $\left[0,1\right]$. Our model was trained as outlined in \texttt{ASGLD}Reg offline training section  with $800$ epochs. We used the Adam optimizer \cite{kingma2014adam} with learning rate of $10^{-3}$ and a batch size of $64$ for training. We added an $L_2$ regularization of network weights and biases to our loss function to avoid overfitting. We empirically set the $L_2$ regularization weight to $10^{-5}$ and the algorithm input parameters to $\alpha=100$ and $\lambda=0.05$. 
%The mean-squared-error (MSE) was exploited as the dissimilarity loss, due to invariability of intensities between different images. 
Figure~\ref{fig:loss}\protect\subref{fig2:a} presents the loss curves as a function of the epoch number, for both the validation and the training sets. The \textit{burn-in} phase ends after $500$ iterations. Hence, we selected $t_b=720$ as the iteration to start the posterior sampling from. A proper selection of $t_b$ has an impact on the resulting registration. Lastly, we calculate the mean and variance of the deformations obtained as output of the CNN in iterations ${t_b,t_{b}+1,...,N}$, and resample the moving image with the averaged deformation to obtain the registered result. The top row of Figure~\ref{fig:regresults} presents an example of a pair of fixed and moving images, the corresponding estimated deformation field, and the resulted warped image from the MNIST database.

\paragraph{MRI Registration}
%We extracted the $10$ central 2D slices which contain a large portion of the brain for each subject of the $10$ subjects of the MGH10 dataset. We split the extracted $100$ 2D images into $64$, $16$ and $20$ examples for training, test and validation, respectively.
We extracted 10 central 2D slices, images which contain large portions of the brain,  from each of the 10 subjects of the MGH10 dataset. We then split the extracted 100 2D images into 64, 16 and 20 examples for training, test and validation, respectively. We generated from these sets $4032$, $240$, $380$ input pairs, respectively. All images were multiplied by their provided brain masks, resized to $128\times 128$ and normalized to the $\left[0,1\right]$ gray-scale domain. Similarly to the MNIST registration above, the network was trained as outlined in \texttt{ASGLD}Reg ($\alpha=80$) with $10,000$ iterations. We used the Adam optimizer \cite{kingma2014adam} with the same setting as for the MNIST registration above, except for the learning rate which was set to $0.5^{-3}$.
%We used the LCC metric was used as the dissimilarity term in the loss function, since contrasts and intensities vary between images. 
The loss curves for both training and validation sets are highlighted in Figure~\ref{fig:loss}\protect\subref{fig2:b}. The training loss curve shows noisy behavior, however, the validation loss curve is more stable after terminating the \textit{burn-in} phase (i.e. in iterations larger than $8000$). We selected $t_b=9800$ and calculated the mean and variance estimates based on deformations sampled from the range $\left[9800,10000\right]$. An example of the deformation estimates obtained with the system and the corresponding registered result is presented in the bottom row of Figure~\ref{fig:regresults}. Additional examples are presented in the supplementary material.       
\paragraph{Evaluation}
We assessed the performance of our Bayesian unsupervised registration system by means of MSE and and Dice score \cite{dice1945measures} for the MNIST and the MRI registration tasks, respectively. The Dice score measures the registration accuracy by quantifying the volume overlap between a reference segmentation and the propagated (registered) structure. The Dice score varies in the range $\left[0,1\right]$, a Dice score of 1 implies identical structures and complete overlap, while a Dice score of 0 implies no overlap between the sets. 
We compared three registration methods: our \texttt{ASGLD}Reg method with the average over the deformations obtained after the iteration $t_b$ (denoted by Averaged), our \texttt{ASGLD}Reg method without averaging (i.e considering only the deformation obtained in the last iteration), and the baseline \texttt{VoxelMorph} \cite{balakrishnan2018unsupervised}.\footnote[5]{\texttt{VoxelMorph} is trained with Adam optimizer with the same settings but without gradients noise injection.}  
We assessed the robustness of our approach against noisy images by corrupting the input images with Gaussian noise with various std.  
%In MRI registration, different methods may yield warped results that are looking similar and one can't distinguish observable differences by the naked-eye only. Therefore, having manually delineated anatomical structures and indicated landmarks is essential to analyze the registration performance quantitatively. 
In our MRI registration experiment on MRI data, we considered the largest $4$ anatomical structures present in the images. The mean and std Dice were calculated over the $4$ anatomical structures. A detailed analysis per each structure is provided in the supplementary materials. 
Table~\ref{tab:results} summarizes the quantitative results obtained by the three methods for the different noise levels. Our \texttt{ASGLD}Reg method shows a statistically significant improvement (paired Student's t-test, $p<0.05$ for MNIST and $p<10^{-5}$ for MGH10) upon the \texttt{VoxelMorph} approach for both the MNIST registration (lower MSE) and the MRI registration (higher Dice score) for all noise levels. The added-value of averaging over the posterior samples rather than taking the weights of the last training iteration is more evident in the noisy scenario. 

\begin{figure}[t]
	\begin{minipage}[b]{0.21\linewidth}
		\centering
		\subfloat[\label{fig4:a}\footnotesize  $A_M$]{
			\includegraphics[width=4.1cm]{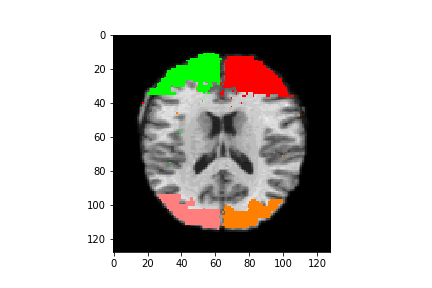}}
	\end{minipage}
	\begin{minipage}[b]{0.21\linewidth}
		\centering
		\subfloat[\label{fig4:b}\footnotesize $A_F$]{
			\includegraphics[width=4.1cm]{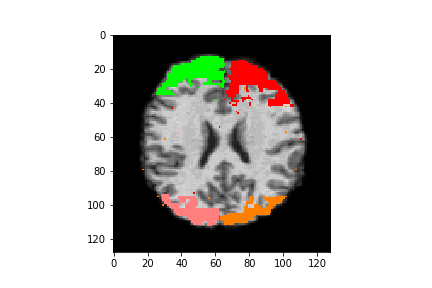}}
	\end{minipage}	
	\begin{minipage}[b]{0.21\linewidth}
		\centering
		\subfloat[\label{fig4:c}\footnotesize The x-component of $\hat{\Phi}$ ]{
			\includegraphics[width=4.1cm]{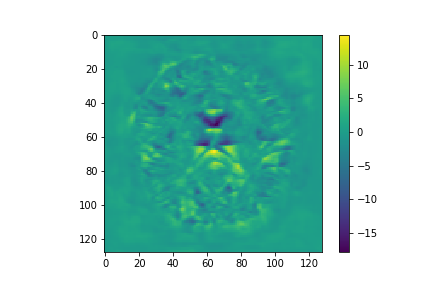}}
	\end{minipage}
	\begin{minipage}[b]{0.21\linewidth}
		\centering
		\subfloat[\label{fig4:d}\footnotesize $A_M\circ \hat{\Phi}$]{
			\includegraphics[width=4.1cm]{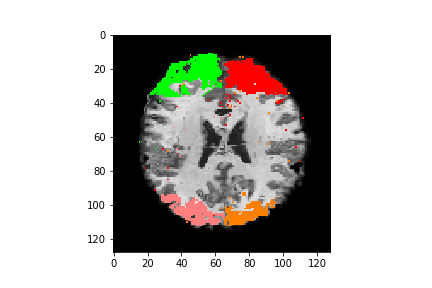}}
	\end{minipage}
	\caption{Registration Results. \protect\subref{fig4:a}, \protect\subref{fig4:b} moving and fixed input images, annotated with the $4$ anatomical segments, respectively. \protect\subref{fig4:c} The x-component of the corresponding deformation field estimate. \protect\subref{fig4:d} The registered image with the propagated structures after warping with $\hat{\Phi}$.   }\label{fig:atlasresults}	
\end{figure} 
\renewcommand{\arraystretch}{1.4}  
\begin{table}
	\centering
	\caption{Registration evaluation results for the MNIST and MRI registration experiments. The added noise level is denoted by $\sigma$. For the MRI registration, mean (std) over the different anatomical structures are presented.} \label{tab:results}
	\resizebox{0.97\linewidth}{!}{
		\begin{tabular}{cc|c|c|c|c|c|c|c|}
			%\hline 
			\toprule
			& \multicolumn{4}{c}{\textbf{MNIST (MSE)}} & \multicolumn{4}{c}{\textbf{MGH10 (Dice)}}\tabularnewline
			%\hline 
			\cmidrule(lr){2-5}
			\cmidrule(lr){6-9}
			& $\sigma=0$ & $\sigma=0.05$ & $\sigma=0.1$ & $\sigma=0.18$ & $\sigma=0$ & $\sigma=0.1$ & $\sigma=0.18$ & $\sigma=0.24$\tabularnewline 
			\hline 
			Averaged & $\textbf{0.0063}$ & $\textbf{0.0077}$ & $\textbf{0.0115}$ & $\textbf{0.0232}$ & $\textbf{0.736 (0.017)} $ & $\textbf{0.732 (0.018)} $ & $\textbf{0.722 (0.018)}$ & $\textbf{0.712 (0.018)} $\tabularnewline 
			\hline 
			Noisy & $0.0064$ & $\textbf{0.0077}$ & $0.0117$ & $0.0235$ & $\textbf{0.736 (0.017)} $ & $0.731 (0.017)$ & $0.721 (0.018)$ & $0.7106 (0.017) $\tabularnewline
			\hline 
			\texttt{VoxelMorph} & $0.0065$ & $0.0079$ & $0.0122$ & $0.0257$ & $0.7109 (0.027)$ & $0.708 (0.026) $ & $0.699 (0.026) $ & $0.691 (0.026) $\tabularnewline
			\bottomrule 
	\end{tabular}}
\end{table}

\section{Discussion and Conclusions}\label{sec:conc}
We proposed a Bayesian unsupervised DL-based registration. We use adaptive noise injection to the training loss gradients to efficiently sample the true posterior distribution of the network weights. Our approach provides empirical estimates of the two principle moments of the deformation field. However, other statistics or higher-order moments can be calculated directly from the posterior samples. Our experiments showed that a Bayesian DL-based registration with posterior sampling through gradient noise injection improves the registration accuracy by means of MSE and Dice score on both the MNSIT and the MGH10 brain MRI databases compared to the point estimates predicted by previously published DL-based registration methods. In our study, we used a 2D version of the \texttt{VoxelMorph} model as the main building-block in which our Bayesian approach method is built upon. However, our proposed Bayesian approach can be extended for 3D registration by replacing the CNN layers from 2D to 3D convolutions. In addition, our model can be extended to ensure diffeomorphic  deformation field estimation by using diffeomorphic integration layers \cite{dalca2019varreg}. Further, the adaptive posterior sampling proposed here can be generalized to other predictive DL models.\\

%\begin{ack}

%\end{ack}

\newpage

%\section*{References}
\medskip
\small
\bibliographystyle{unsrt}
\bibliography{refs}{}

\end{document}